\documentclass[journal]{IEEEtran}
\usepackage{amsmath}
\usepackage{algorithm}
\usepackage{algorithmic}
\usepackage{graphicx}
\usepackage{multirow}
\usepackage{tabularx}
 
\usepackage{soul}
\usepackage{colortbl}  
\usepackage{color,xcolor}
\usepackage{array}
\usepackage{amssymb}
\usepackage{bbding}
\usepackage{subfigure}
\usepackage{diagbox}
\usepackage{rotating}
\usepackage{bm}
\usepackage{tikz} 
\usepackage{booktabs}
\usepackage{cite}
\usepackage{CJK}
\usepackage{overpic}
\usepackage[colorlinks,
            linkcolor=red,
            anchorcolor=blue,
            citecolor=black
            ]{hyperref}

\newcommand{\addFig}[1]{}
\newcommand{\addFigs}[1]{}

\usepackage{subfigure}

\newcommand{\ie}{\textit{i}.\textit{e}.,~}

\definecolor{lightgreen}{HTML}{9aff99}
\definecolor{lightpink}{HTML}{FFE2E1}
\definecolor{lightblue}{HTML}{c9e5ff}
\definecolor{lightpurple}{HTML}{abacf4}
\definecolor{lightorange}{HTML}{ffcc67}
\definecolor{lightgray}{HTML}{c0c0c0}
\definecolor{lightred}{HTML}{FF8181}

\soulregister{\cite}7 
\soulregister{\citep}7 
\soulregister{\citet}7 
\soulregister{\ref}7 
\soulregister{\pageref}7 

\usepackage{balance}
\usepackage{cleveref}
\crefformat{section}{\S#2#1#3} 
\crefformat{subsection}{\S#2#1#3}
\crefformat{subsubsection}{\S#2#1#3}

\ifCLASSINFOpdf
\else
\fi

\begin{document}
%
\title{Texture-Semantic Collaboration Network for \\ORSI Salient Object Detection}
%
%
%

\author{Gongyang~Li,
	Zhen~Bai,
	and~Zhi~Liu,~\IEEEmembership{Senior Member,~IEEE}
        
\thanks{The authors are with Key Laboratory of Specialty Fiber Optics and Optical Access Networks, Joint International Research Laboratory of Specialty Fiber Optics and Advanced Communication, Shanghai Institute for Advanced Communication and Data Science, Shanghai University, Shanghai 200444, China, and School of Communication and Information Engineering, Shanghai University, Shanghai 200444, China. Gongyang Li and Zhi Liu are also with Wenzhou Institute of Shanghai University, Wenzhou 325000, China (email: ligongyang@shu.edu.cn; bz536476@163.com; liuzhisjtu@163.com).}
\thanks{\textit{Corresponding author: Zhi Liu.}}
}

\markboth{IEEE TRANSACTIONS ON CIRCUITS AND SYSTEMS II: EXPRESS BRIEFS}
{Shell \MakeLowercase{\textit{et al.}}: Bare Demo of IEEEtran.cls for IEEE Journals}

\maketitle

\begin{abstract}
Salient object detection (SOD) in optical remote sensing images (ORSIs) has become increasingly popular recently.
Due to the characteristics of ORSIs, ORSI-SOD is full of challenges, such as multiple objects, small objects, low illuminations, and irregular shapes.
To address these challenges, we propose a concise yet effective \emph{Texture-Semantic Collaboration Network} (TSCNet) to explore the collaboration of texture cues and semantic cues for ORSI-SOD.
Specifically, TSCNet is based on the generic encoder-decoder structure.
In addition to the encoder and decoder, TSCNet includes a vital Texture-Semantic Collaboration Module (TSCM), which performs valuable feature modulation and interaction on basic features extracted from the encoder.
The main idea of our TSCM is to make full use of the texture features at the lowest level and the semantic features at the highest level to achieve the expression enhancement of salient regions on features.
In the TSCM, we first enhance the position of potential salient regions using semantic features.
Then, we render and restore the object details using the texture features.
Meanwhile, we also perceive regions of various scales, and construct interactions between different regions.
Thanks to the perfect combination of TSCM and generic structure, our TSCNet can take care of both the position and details of salient objects, effectively handling various scenes.
Extensive experiments on three datasets demonstrate that our TSCNet achieves competitive performance compared to 14 state-of-the-art methods.
The code and results of our method are available at https://github.com/MathLee/TSCNet.
\end{abstract}


\IEEEpeerreviewmaketitle

\section{Introduction}
\IEEEPARstart{S}{alient} object detection (SOD) focuses on extracting the most attractive objects/regions in a scene~\cite{R1,R2,R3}. 
Recently, SOD in optical remote sensing images (ORSIs) has become a shining topic in the SOD community, and aims to capture the most attention-grabbing ships, airplanes, cars, buildings, islands, rivers, \textit{etc.}, in ORSIs.
ORSI-SOD has great applications in urban planning, land resource evaluation, environmental monitoring, and agricultural production~\cite{2019LVNet,2021DAFNet,2021RRNet,2023GeleNet}.

Researchers have made remarkable achievements in SOD in natural scene images (NSIs)~\cite{2021SUCA,2021PAKRN,2021VST,2017Amulet,19HRSOD,21DRFNet,23RMFormer}.
However, due to the differences in scenes and shooting between NSIs and ORSIs, NSI-SOD methods may not always be able to handle challenging scenes in ORSIs well.
Therefore, many efforts have been made in ORSI-SOD, resulting in some effective specialized solutions.
Among existing specialized ORSI-SOD methods, some methods focus on mining informative clues from features at a single level~\cite{2022MCCNet,2022ERPNet} (called the single-level type), while some focus on extracting contextual clues from features at adjacent levels~\cite{2019LVNet,2022CorrNet,2022HFANet,2022ACCoNet} (called the adjacent-level type).
Differently, some methods focus on the details of salient objects, and explore the edge information for ORSI-SOD~\cite{2022EMFINet,2022MJRBM} (called the detail type).
While some focus on the position of salient objects, and explore the global semantic information for ORSI-SOD~\cite{2022GPNet,2023SeaNet} (called the position type).
Although these four types of methods have promoted the development of ORSI-SOD, each type of method has its own drawbacks.
Obviously, the single-level type ignores contextual information.
The adjacent-level type only utilizes information from nearby adjacent levels, ignoring information from longer distances (\ie levels).
The detail type and the position type only consider edge information or position information, and both types of methods are suboptimal.

Inspired by the above observations, we integrate the advantages of four types of ORSI-SOD methods to alleviate their problems.
Concretely, we utilize the contextual information from further distances (\ie levels), and take into account both detail and position information.
We believe that the above information is essential to handle various complex and ever-changing scenes in ORSIs.
Based on this idea, we propose a concise yet effective Texture-Semantic Collaboration Network (TSCNet) for ORSI-SOD, which makes an attempt to explore the collaboration of texture cues and semantic cues.
Similar to previous ORSI-SOD methods, our TSCNet also follows the classic encoder-decoder structure~\cite{2022TSCII}.
In TSCNet, we propose the key Texture-Semantic Collaboration Module (TSCM) to modulate current features using semantic features and texture features, enabling current features to obtain valuable position and detail information.
We also introduce the transformer~\cite{2021ViT} into TSCM to establish region-level connections, which is effective in handling scenes of multiple objects.
In this way, our TSCNet can handle various challenging scenes of ORSIs.
The saliency map generated by our TSCNet can accurately locate salient objects and finely outline the details of salient objects, making our TSCNet a competitive detector.

Our main contributions are threefold:
\begin{itemize}
\item We explore the collaboration of texture cues and semantic cues for ORSI-SOD, and propose a novel \emph{TSCNet}, which takes into account both position highlighting and detail rendering of salient objects in ORSIs.

\item We propose a \emph{Texture-Semantic Collaboration Module} to coordinate the lowest-level texture features and the highest-level semantic features to perform valuable modulation and interaction on other levels of features, enhancing the expression of salient regions.

\item We evaluate the proposed TSCNet on three public ORSI-SOD datasets, \ie EORSSD, ORSSD, and ORSI-4199.
Comprehensive experiments show that our TSCNet is competitive and that our key module is effective.

\end{itemize}

\begin{figure}[t!]
	\centering
	\begin{overpic}[width=1\columnwidth]{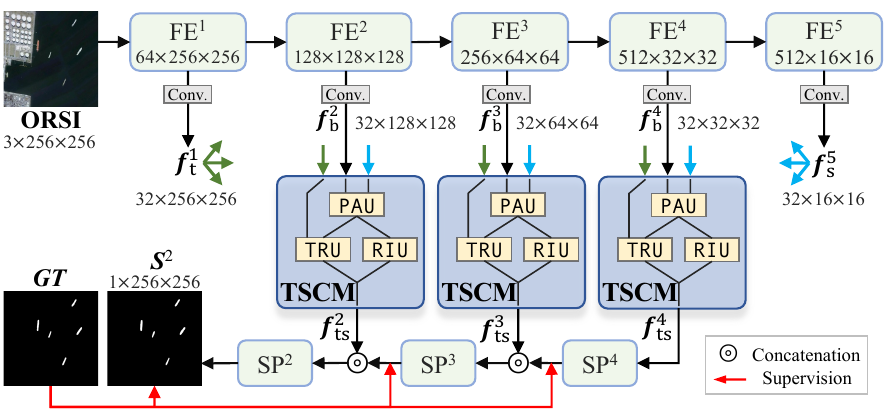}
    \end{overpic}
	\caption{Architecture of the proposed TSCNet.
    }
    \label{fig:Framework}
\end{figure}

\section{Proposed Method}
\label{sec:OurMethod}
%


\subsection{Network Overview}
\label{sec:Overview}
As illustrated in Fig.~\ref{fig:Framework}, the proposed TSCNet is based on the encoder-decoder structure, and consists of the encoder, the Texture-Semantic Collaboration Module (TSCM), and the decoder.
We adopt the VGG~\cite{2014VGG16ICLR} as our encoder with the input size of $3\!\times\!256\!\times\!256$, also known as the feature extractor (FE), and denote its basic block as FE$^{i}$ ($i=1,2,3,4,5$).
We use a convolution layer after each basic block to compress the channel number of features, resulting in five-level features.
This operation can significantly reduce computational complexity.
Here, we denote the features of FE$^{1}$ as $\boldsymbol{f}^{1}_{\rm t} \in \mathbb{R}^{32\!\times\!256\!\times\!256}$ (\ie texture features), the features of FE$^{5}$ as $\boldsymbol{f}^{5}_{\rm s} \in \mathbb{R}^{32\!\times\!16\!\times\!16}$ (\ie semantic features), and the other features as $\boldsymbol{f}^{i}_{\rm b} \in \mathbb{R}^{c_i\!\times\!h_i\!\times\!w_i}$ ($i=2,3,4$), where $h_i$ and $w_i$ are $\frac{256}{2^{i-1}}$, and $c_i$ is 32.
Then, we transfer the valuable cues of $\boldsymbol{f}^{1}_{\rm t}$ and $\boldsymbol{f}^{5}_{\rm s}$ to $\boldsymbol{f}^{i}_{\rm b}$ using the TSCM.
In the TSCM, we first adopt Position Anchoring Unit (PAU) to explore the semantic features to enhance the position of salient regions in both channel and spatial dimensions.
Next, we adopt Texture Rendering Unit (TRU) to explore the texture features to render and restore the object details, and adopt Region Interaction Unit (RIU) to construct interactions between different regions.
In this way, we get $\boldsymbol{f}^{i}_{\rm ts} \in \mathbb{R}^{c_i\!\times\!2h_i\!\times\!2w_i}$ ($i=2,3,4$) from TSCM.
Finally, we gradually infer the saliency map $\boldsymbol{S}^{2}\in \mathbb{R}^{1\!\times\!256\!\times\!256}$ in the decoder using the saliency prediction block denoted as SP$^{i}$ ($i=2,3,4$).
Notably, the deep supervision is introduced in the training phase to accelerate network convergence.

\begin{figure}[t!]
	\centering
	\begin{overpic}[width=0.9\columnwidth]{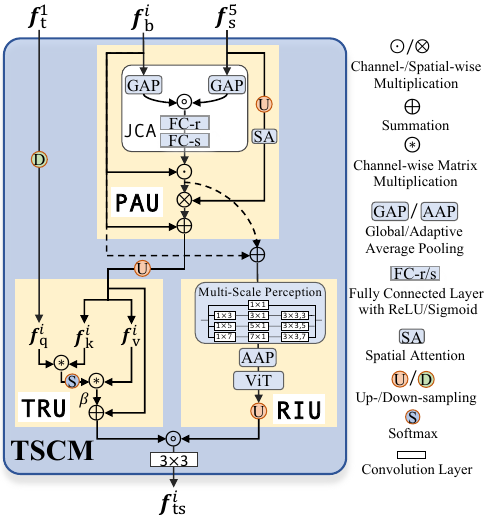}
    \end{overpic}
	\caption{Illustration of the TSCM, which consists of PAU, TRU, and RIU.
    }
    \label{fig:TSCM}
\end{figure}

\subsection{Texture-Semantic Collaboration Module}
\label{sec:TSCM}
As we all know, in convolutional neural networks, as the number of convolution layers increases, the texture information of objects will gradually be lost, and the semantic information (\ie position information) will dominate.
In other words, the low-level features contain detailed texture cues, while the high-level features contain sufficient semantic cues.
Due to the complexity and variability of ORSIs, utilizing texture cues and semantic cues simultaneously is beneficial for ORSI-SOD.
Therefore, we propose TSCM to coordinate the lowest-level texture features, \ie $\boldsymbol{f}^{1}_{\rm t}$, and the highest-level semantic features, \ie $\boldsymbol{f}^{5}_{\rm s}$, to perform valuable modulation and interaction on $\boldsymbol{f}^{i}_{\rm b}$.
We illustrate the detailed structure of TSCM in Fig.~\ref{fig:TSCM}, which consists of PAU, TRU, and RIU.

\textit{1) Position Anchoring Unit}.
As shown at the top of Fig.~\ref{fig:TSCM}, the inputs of PAU are $\boldsymbol{f}^{i}_{\rm b}$ and $\boldsymbol{f}^{5}_{\rm s}$.
Obviously, in PAU, we aim at exploring the semantic information of $\boldsymbol{f}^{5}_{\rm s}$ to anchor salient objects in both channel and spatial dimensions using the attention mechanism~\cite{2018CBAM}.

Differently from traditional channel attention (CA), we utilize $\boldsymbol{f}^{5}_{\rm s}$ to assist $\boldsymbol{f}^{i}_{\rm b}$ to achieve accurate channel-wise feature enhancement, which we call joint channel attention (JCA).
We first extremely compress $\boldsymbol{f}^{i}_{\rm b}$ and $\boldsymbol{f}^{5}_{\rm s}$ to the global representation with the size of $32\times1\times1$.
Then, we concatenate them and adopt two fully connected layers with suitable activation functions to generate the joint channel attention map with the size of $32\times1\times1$.
Next, similar to traditional CA, we modulate $\boldsymbol{f}^{i}_{\rm b}$ using the joint channel attention map to highlight important channels, generating $\boldsymbol{f}^{i}_{\rm jca} \in \mathbb{R}^{c_i\!\times\!h_i\!\times\!w_i}$.
We formulate JCA as follows:
\begin{equation}
   \begin{aligned}
    \boldsymbol{f}^{i}_{\rm jca} =  {\rm FC_{s}}\Big({\rm FC_{r}} \big( {\rm GAP} (\boldsymbol{f}^{i}_{\rm b}) \circledcirc {\rm GAP} (\boldsymbol{f}^{5}_{\rm s})\big) \Big) \odot \boldsymbol{f}^{i}_{\rm b},
    \label{eq:JCA}
    \end{aligned}
\end{equation}
where $\circledcirc$ is the concatenation operator, $\odot$ is the channel-wise multiplication, ${\rm GAP}(\cdot)$ is the global average pooling layer, and ${\rm FC_{r}}(\cdot)$/${\rm FC_{s}}(\cdot)$ is the fully connected layer with ReLU/sigmoid activation function.
With the help of semantic cues, our JCA is more sensitive to informative channels than traditional CA.

At the rest of PAU, we focus on the spatial enhancement of $\boldsymbol{f}^{i}_{\rm jca}$ to anchor the position of salient objects, which is a supplement to JCA.
We extract the spatial attention map from $\boldsymbol{f}^{5}_{\rm s}$, and anchor salient regions of $\boldsymbol{f}^{i}_{\rm jca}$ through the spatial-wise multiplication.
Moreover, we adopt the residual connection to fuse $\boldsymbol{f}^{i}_{\rm b}$ and the effectively enhanced features, generating the output features of PAU, \ie $\boldsymbol{f}^{i}_{\rm pau} \in \mathbb{R}^{c_i\!\times\!h_i\!\times\!w_i}$.

\textit{2) Texture Rendering Unit}.
The position of salient objects is well highlighted in the PAU.
We turn our attention to the texture of salient objects.
In TRU, we aim to achieve the feature super-resolution of $\boldsymbol{f}^{i}_{\rm pau}$ with the assistance of $\boldsymbol{f}^{1}_{\rm t}$, that is, we not only restore the texture of $\boldsymbol{f}^{i}_{\rm pau}$, but also enlarge the size of $\boldsymbol{f}^{i}_{\rm pau}$.

We first up-sample $\boldsymbol{f}^{i}_{\rm pau}$ from $32\!\times\!h_i\!\times\!w_i$ to $32\!\times\!2h_i\!\times\!2w_i$, generating $\boldsymbol{\hat{f}}^{i}_{\rm pau}$.
Meanwhile, we down-sample $\boldsymbol{f}^{1}_{\rm t}$ from $32\!\times\!256\!\times\!256$ to $32\!\times\!2h_i\!\times\!2w_i$, generating $\boldsymbol{\hat{f}}^{1}_{\rm t}$.
Next, according to the standard self-attention mechanism~\cite{2017transformer}, we generate the corresponding $\boldsymbol{f}^{i}_{\rm q}$ and $\{\boldsymbol{f}^{i}_{\rm k},\boldsymbol{f}^{i}_{\rm v}\}$ from $\boldsymbol{\hat{f}}^{1}_{\rm t}$ and $\boldsymbol{\hat{f}}^{i}_{\rm pau}$, respectively.
We model correlations between $\boldsymbol{f}^{i}_{\rm q}$ and $\boldsymbol{f}^{i}_{\rm k}$, and transfer them to $\boldsymbol{f}^{i}_{\rm v}$ to render the texture of features, generating $\boldsymbol{f}^{i}_{\rm sp}$ with the size of $2h_i\!\times\!2w_i$.
Finally, we also adopt the residual connection to fuse $\boldsymbol{f}^{i}_{\rm sp}$ and $\boldsymbol{\hat{f}}^{i}_{\rm pau}$ with a coefficient of $\beta$, generating the output features of TRU, \ie $\boldsymbol{f}^{i}_{\rm tru} \in \mathbb{R}^{c_i\!\times\!2h_i\!\times\!2w_i}$.

Notably, the matrix multiplication used in our TRU is the channel-wise matrix multiplication, which is different from that in the standard self-attention mechanism.
Compared to the standard one, our channel-wise matrix multiplication can significantly reduce computational complexity and memory usage, enabling the use of self-attention attention at multiple levels.
As is well known, for $\{\boldsymbol{f}^{1},\boldsymbol{f}^{2}\}\!\in\!\mathbb{R}^{c\times\!h\times\!w}$, the  standard one performs matrix multiplication at the size of $(hw\times\!c)\times\!(c\times\!hw)$, generating features with size of $hw\times\!hw$.
Differently, our TRU only performs the matrix multiplication for two corresponding channels at the size of $(h\times\!w)\times\!(w\times\!h)$, and there are $c$ channels in total, generating features with the size of $c\times\!(h\times\!w)$\footnote{In our TSCNet, since $h$ is equal to $w$, $h\times\!h$ is the same as $h\times\!w$.}.

\textit{3) Region Interaction Unit}.
Similar to TRU, RIU can also perform the function of rendering textures, but it reconstructs textures by establishing region-level connections between different regions.
The input of RIU is the summation of $\boldsymbol{f}^{i}_{\rm jca}$ and $\boldsymbol{f}^{i}_{\rm b}$, denoted as $\boldsymbol{f}^{i}_{\rm in}\!\in\!\mathbb{R}^{c_i\!\times\!h_i\!\times\!w_i}$.
According to the fact that salient objects in ORSIs have various sizes and shapes, we first adopt the multi-scale perception operation to perceive regions of various scales.
The multi-scale perception operation has four parallel dilated convolution branches\cite{Dila2016}, which can be formulated as follows:
\begin{equation}
   \begin{aligned}
    \boldsymbol{f}^{i,j}_{\rm br}=\left\{
	\begin{array}{lll}
	{\rm C_{1\times 1}} ( \boldsymbol{f}^{i}_{\rm in} ) ,     & j=1,\\
	{\rm C_{3\times 3, k}} ({\rm C_{k\times 1}} ({\rm C_{1\times k}} (\boldsymbol{f}^{i}_{\rm in}))) ,      & j=2,3,4; k=2j-1, \\
	\end{array}  \right. 
    \label{eq:MSP}
    \end{aligned}
\end{equation}
where $\boldsymbol{f}^{i,j}_{\rm br} \!\in\!\mathbb{R}^{c_i\!\times\!h_i\!\times\!w_i}$ is the output of $j$-th branch and ${\rm C_{k_1\times k_2, r}}(\cdot)$ is the convolution layer with kernel size of $\rm k_1\times k_2$ and dilated rate of $\rm r$.
The output of these four branches is fused through the concatenation and a convolution layer, generating $\boldsymbol{f}^{i}_{\rm msp}\!\in\!\mathbb{R}^{c_i\!\times\!h_i\!\times\!w_i}$.

Moreover, we adopt the effective transformer (\ie ViT\cite{2021ViT}) to comprehensively model the long-range texture dependencies of different regions through the multi-head self-attention mechanism.
Notably, we set the input size of ViT as $32\!\times\!32\!\times\!32$ and the patch size as $1\!\times\!1$.
This is because that $\boldsymbol{f}^{4}_{\rm msp}$, $\boldsymbol{f}^{3}_{\rm msp}$, and $\boldsymbol{f}^{2}_{\rm msp}$ are with various sizes, if we set the input size and patch size of ViT to the fixed size, we can reconstruct texture at different levels on the same size, thus maintaining texture consistency.
Therefore, as depicted in RIU of Fig.~\ref{fig:TSCM}, we insert an adaptive average pooling layer between the multi-scale perception and ViT to compress $\boldsymbol{f}^{i}_{\rm msp}$ from $h_i\!\times\!w_i$ to $32\!\times\!32$.
Then, we enhance the region interaction using ViT, and up-sample the output of ViT to $c_i\!\times\!2h_i\!\times\!2w_i$, generating the output features of RIU, \ie $\boldsymbol{f}^{i}_{\rm riu} \in \mathbb{R}^{c_i\!\times\!2h_i\!\times\!2w_i}$.

Finally, we adopt the concatenation operator and the convolution layer to fuse $\boldsymbol{f}^{i}_{\rm tru}$ and $\boldsymbol{f}^{i}_{\rm riu}$, generating the output features of TSCM, \ie $\boldsymbol{f}^{i}_{\rm ts} \in \mathbb{R}^{c_i\!\times\!2h_i\!\times\!2w_i}$.
Through the collaboration of these three units, our TSCM makes full use of texture and semantic features to enhance the expression of salient regions in a comprehensive manner.

\begin{table*}[t!]
  \centering
  \footnotesize
  \renewcommand{\arraystretch}{0.8}
  \renewcommand{\tabcolsep}{1.7mm}
  \caption{
   Quantitative comparisons with state-of-the-art methods on EORSSD, ORSSD, and ORSI-4199 datasets.
   $\uparrow$ indicates that the higher the better, while $\downarrow$ is the opposite.
   We mark the results that are better than our method in \textcolor{blue}{\textbf{blue}}.
    }
\label{table:QuantitativeResults}
  
\begin{tabular}{r|cccc|cccc|cccc}
\midrule[1pt]    
 \multirow{2}{*}{Methods}
 & \multicolumn{4}{c|}{EORSSD~\cite{2021DAFNet}} 
 & \multicolumn{4}{c|}{ORSSD~\cite{2019LVNet}} 
 & \multicolumn{4}{c}{ORSI-4199~\cite{2022MJRBM}}   \\
 
 \cline{2-5} \cline{6-9} \cline{10-13} 
       & $S_{\alpha}\uparrow$ & $F_{\beta}^{\rm{mean}}\uparrow$ & $E_{\xi}^{\rm{mean}}\uparrow$ & $ \mathcal{M}\downarrow$
   	          & $S_{\alpha}\uparrow$ & $F_{\beta}^{\rm{mean}}\uparrow$ & $E_{\xi}^{\rm{mean}}\uparrow$ & $ \mathcal{M}\downarrow$
	          & $S_{\alpha}\uparrow$ & $F_{\beta}^{\rm{mean}}\uparrow$ & $E_{\xi}^{\rm{mean}}\uparrow$ & $ \mathcal{M}\downarrow$  \\
	     
\hline
									   								   
SUCA~\cite{2021SUCA}  	& 0.8988 & 0.7949 & 0.9277 & 0.0097
									   & 0.8989 & 0.8237 & 0.9400 & 0.0145 
									   & \textcolor{blue}{\textbf{0.8794}} & 0.8590 & 0.9356 & 0.0304 \\
									   
PA-KRN~\cite{2021PAKRN}  & 0.9192 & 0.8358 & 0.9536 & 0.0104
									   & 0.9239 & 0.8727 & 0.9620 & 0.0139 
									   & 0.8491 & 0.8324 & 0.9168 & 0.0382 \\
									   								   
VST~\cite{2021VST}        & 0.9208 & 0.8263 & 0.9442 & 0.0067
									   & 0.9365 & 0.8817 & 0.9621 & 0.0094 
									   & \textcolor{blue}{\textbf{0.8790}} & 0.8524 & 0.9348 & \textcolor{blue}{\textbf{0.0281}} \\
\hline
LVNet~\cite{2019LVNet}  	  & 0.8630 & 0.7328 & 0.8801 & 0.0146 
									      & 0.8815 & 0.7995 & 0.9259 & 0.0207
									      & - & - & - & -  \\
									      
DAFNet~\cite{2021DAFNet}   & 0.9166 & 0.7845 & 0.9291 & \textcolor{blue}{\textbf{0.0060}}
									      & 0.9191 & 0.8511 & 0.9539 & 0.0113 
									      & - & - & - & -  \\									     
									   						     
MJRBM~\cite{2022MJRBM}  & 0.9197 & 0.8239 & 0.9350 & 0.0099
									   & 0.9204 & 0.8566 & 0.9415 & 0.0163  
									   & 0.8593 & 0.8309 & 0.9102 & 0.0374 \\
									   
EMFINet~\cite{2022EMFINet} & 0.9290 & 0.8486 & 0.9604 & 0.0084
									     & 0.9366 & 0.8856 & 0.9671 & 0.0109  
									     & 0.8675 & 0.8479 & 0.9257 & 0.0330  \\
									     
ERPNet~\cite{2022ERPNet}  & 0.9210  & 0.8304 & 0.9401 & 0.0089 
									   & 0.9254  & 0.8745 & 0.9566  & 0.0135  
									   & 0.8670 & 0.8374 & 0.9149 & 0.0357 \\
									   
ACCoNet~\cite{2022ACCoNet} 	  & 0.9290 & 0.8552 & 0.9653 & 0.0074
									   & \textcolor{blue}{\textbf{0.9437}} & 0.8971 & 0.9754 & 0.0088 
									   & 0.8775 & 0.8620 & 0.9342 & 0.0314 \\
									   
CorrNet~\cite{2022CorrNet} 		& 0.9289 & 0.8620 & 0.9646 & 0.0083
									   &  0.9380 & 0.9002 & 0.9746 & 0.0098  
									   & 0.8623 & 0.8513 & 0.9206 & 0.0366  \\	
									   
MCCNet~\cite{2022MCCNet} 	  & 0.9327 & 0.8604 & 0.9685 & 0.0066
				       					  & \textcolor{blue}{\textbf{0.9437}} & \textcolor{blue}{\textbf{0.9054}} & 0.9758 & 0.0087 
									  & 0.8746 & 0.8630 & 0.9348 & 0.0316 \\								    	
									   
GPNet~\cite{2022GPNet} 				 & 0.9233 & 0.8447 & 0.9617 & 0.0085
									& 0.9185 & 0.8683 & 0.9590 & 0.0125
									& 0.8573  & 0.8396 & 0.9184 & 0.0384 \\
									   
HFANet~\cite{2022HFANet} 				 & 0.9380 & 0.8681 & 0.9679 & 0.0070
									& 0.9399 & 0.8981 & 0.9712 & 0.0092  
									& 0.8767 & 0.8624 & 0.9336 & 0.0314 \\	
									
SeaNet~\cite{2023SeaNet}				 & 0.9208 & 0.8519 & 0.9651 & 0.0073 
									   & 0.9260 & 0.8772 & 0.9722 & 0.0105 
									   & 0.8722 & 0.8591 & 0.9363 & 0.0308  \\			   	
\hline
\hline

\textbf{TSCNet (Ours)}			& \textbf{0.9383} & \textbf{0.8740} & \textbf{0.9717} & \textbf{0.0061} 
									 & \textbf{0.9428} & \textbf{0.9030} & \textbf{0.9804} & \textbf{0.0081}  
									 &  \textbf{0.8783} & \textbf{0.8703} & \textbf{0.9418} & \textbf{0.0295}  \\	
									   								   								   
\toprule[1pt]

\end{tabular}
\end{table*}
\begin{figure*}[t!]
    \centering
    \footnotesize
	\begin{overpic}[width=1\textwidth]{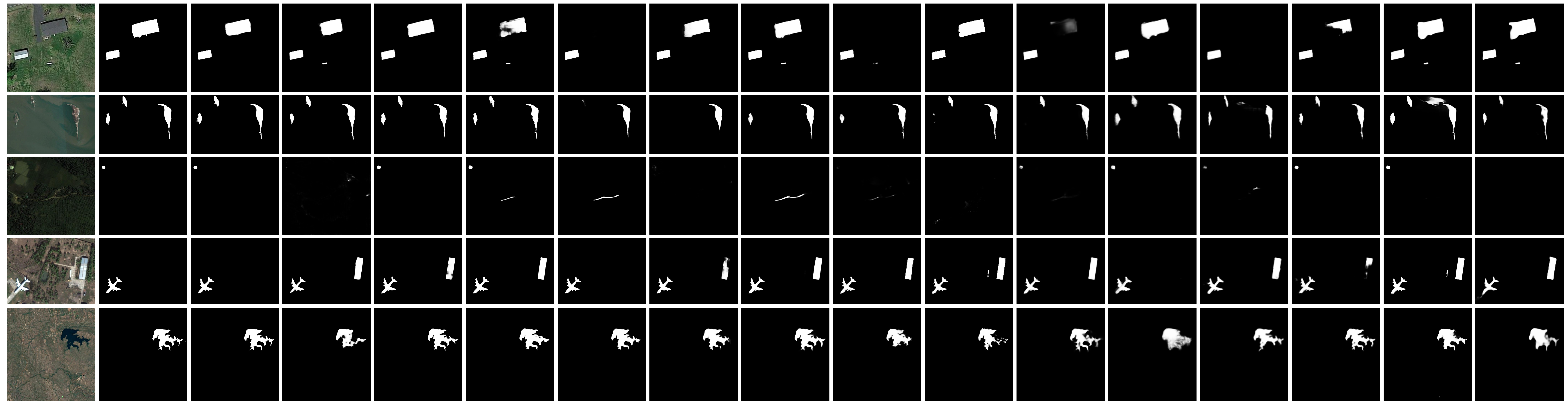}

    \put(0.9,-1.1){ ORSI }
    \put(7.45,-1.1){ GT}
    \put(12.7,-1.1){ \textbf{Ours}}
    \put(18.,-1.1){ SeaNet  }
    \put(23.5,-1.1){ HFANet  }
    \put(29.84,-1.1){ GPNet  }
    \put(34.95,-1.1){ MCCNet }
    \put(41.1,-1.1){ CorrNet }
    \put(46.5,-1.1){ ACCoNet }
    \put(52.83,-1.1){ ERPNet }
    \put(58.28,-1.1){ EFMINet }
    \put(64.4,-1.1){ MJRBM }
    \put(70.25,-1.1){ DAFNet }
    \put(76.7,-1.1){ LVNet }
    \put(83.1,-1.1){ VST }
    \put(87.65,-1.1){ PA-KRN }
    \put(94.3,-1.1){ SUCA } 
    
    \end{overpic}
	\caption{Qualitative comparison with 14 state-of-the-art methods on representative scenes in ORSIs.
    }
    \label{fig:VisualExample}
\end{figure*}
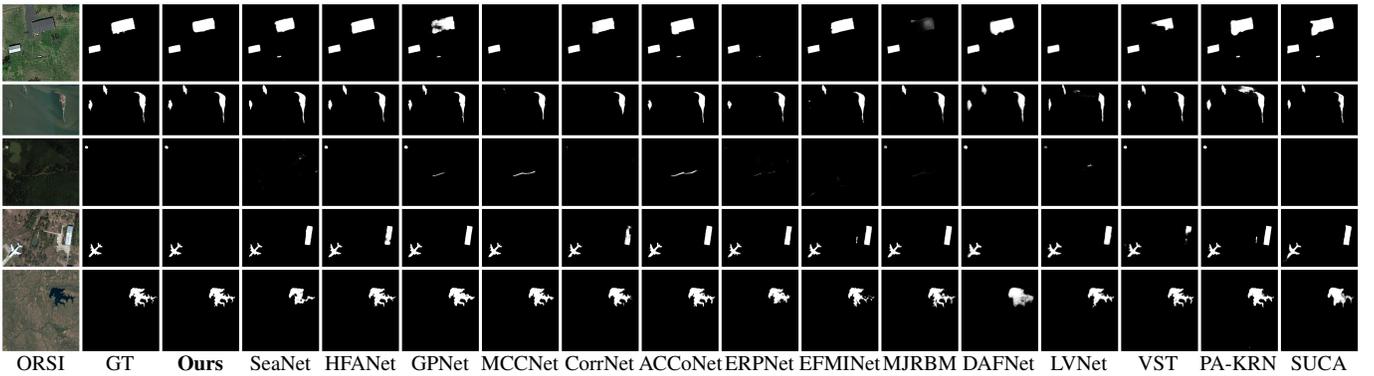

\subsection{Loss Function}
\label{sec:LossFunction}
The basic SP block of our encoder generally consists of two convolution layers, a dropout layer, and a deconvolution layer, while the last SP block (\ie SP$^2$) consists of three convolution layers.
Due to the introduction of deep supervision into the network training, in addition to the last SP block generating the final saliency map $\boldsymbol{S}^{2}$, the other two SP blocks also generate two lateral saliency maps, \ie $\boldsymbol{S}^{3}\in \mathbb{R}^{1\!\times\!256\!\times\!256}$ and $\boldsymbol{S}^{4}\in \mathbb{R}^{1\!\times\!128\!\times\!128}$.
For these three saliency maps, we impose the hybrid loss function, including the binary cross-entropy (BCE) loss and intersection-over-union (IoU) loss, to each one.
Thus, we formulate the total loss function ${L}_{\rm total}$ as follows:
\begin{equation}
   \begin{aligned}
    {L}_{\mathrm{total}} =  \sum_{i = 2}^{4}  \big( \ell_{\mathrm{bce}}^{i} ( \mathrm{up}(\boldsymbol{S}^{i}), \boldsymbol{GT}) + \ell_{\mathrm{iou}}^{i} (\mathrm{up}(\boldsymbol{S}^{i}), \boldsymbol{GT}) \big),
    \label{eq:Loss2}
    \end{aligned}
\end{equation}
where $\ell_{\mathrm{bce}}^i (\cdot)$ and $\ell_{\mathrm{iou}}^i (\cdot)$ are BCE loss and IoU loss, respectively, $\boldsymbol{GT}\in \{0,1\}^{1\!\times\!256\!\times\!256}$ is the binary ground truth (GT) map, and $\mathrm{up}(\cdot)$ is the up-sampling operation if the sizes of $\boldsymbol{S}^{i}$ and $\boldsymbol{GT}$ do not match.

\section{Experiments}
\label{sec:exp}

\subsection{Experimental Setup}
\label{sec:ExpProtocol}
\textit{1) Datasets.}
We evaluate our TSCNet on three datasets, \ie ORSSD~\cite{2019LVNet}, EORSSD~\cite{2021DAFNet}, and ORSI-4199~\cite{2022MJRBM}.
ORSSD has 800 images, of which 600 images are for training and 200 images are for testing.
EORSSD has 2000 images, of which 1400 images are for training and 600 images are for testing.
ORSI-4199 has 4199 images, of which 2000 images are for training and 2199 images are for testing.

\textit{2) Evaluation Metrics.}
We use four quantitative evaluation metrics for performance measurement, including
S-measure ($S_{\alpha}$, $\alpha$ = 0.5)~\cite{Fan2017Smeasure},
mean F-measure ($F_{\beta}$, $\beta^{2}$ = 0.3)~\cite{Fmeasure},
mean E-measure ($E_{\xi}$)~\cite{Fan2018Emeasure}, and
mean absolute error ($\mathcal{M}$).

\textit{3) Implementation Details.}
We conduct all experiments using the PyTorch on a computer with an NVIDIA RTX 3090 GPU (24GB memory).
During the training phase, all images and ground truths are resized to 256$\times$256, and then flipping and rotation are adopted for data augmentation.
We use the Adam optimizer for parameter updating with a base learning rate of $1e^{-4}$ and a batch size of 4.
The learning rate drops to 1/10 every 30 epochs.
On each dataset, we train our TSCNet for 70 epochs on its training set, and then test the trained TSCNet on its testing set.

\subsection{Comparison with State-of-the-arts}
We compare our TSCNet with 14 state-of-the-art SOD methods for NSIs and ORSIs.
They are SUCA~\cite{2021SUCA}, PA-KRN~\cite{2021PAKRN} and VST~\cite{2021VST} used for NSI-SOD,
and LVNet~\cite{2019LVNet}, DAFNet~\cite{2021DAFNet}, MJRBM~\cite{2022MJRBM}, EMFINet~\cite{2022EMFINet}, ERPNet~\cite{2022ERPNet}, ACCoNet~\cite{2022ACCoNet}, CorrNet~\cite{2022CorrNet}, MCCNet~\cite{2022MCCNet}, GPNet~\cite{2022GPNet}, HFANet~\cite{2022HFANet} and SeaNet~\cite{2023SeaNet} used for ORSI-SOD.
%

We report the quantitative performance of all methods in Tab.~\ref{table:QuantitativeResults}.
Among all 12 evaluation metrics, our method outperforms all compared methods in seven evaluation metrics, ranks second on four evaluation metrics, and ranks third on one evaluation metric.
%
Moreover, we show the qualitative comparison on three challenging ORSI scenes in Fig.~\ref{fig:VisualExample}, including multiple objects, small objects, and irregular shapes.
We can clearly observe that our method handles these challenging scenes well, and the saliency maps generated by our method are more accurate and complete than those generated by other competitors.
The above quantitative and qualitative comparisons indicate that our method is a competitive saliency detector for ORSIs.

\begin{table}[!t]
\centering
\caption{Ablation results of evaluating the contribution of each component in TSCNet.
  The best one in each column is \textbf{bold}.
  }
\label{Ablation_component}
\renewcommand{\arraystretch}{0.8}
\renewcommand{\tabcolsep}{1.mm}
\begin{tabular}{c|c|ccc||cccc}
\bottomrule

 \multirow{2}{*}{No.} & \multirow{2}{*}{Baseline} 
 & \multicolumn{3}{c||}{TSCM}  & \multicolumn{4}{c}{EORSSD~\cite{2021DAFNet}}  \\
 
 \cline{6-9}
    & & PAU & TRU & RIU 
    & $S_{\alpha}\uparrow$ & $F_{\beta}^{\rm{mean}}\uparrow$ & $E_{\xi}^{\rm{mean}}\uparrow$ & $ \mathcal{M}\downarrow$ \\
\hline
\hline
1 &  \Checkmark &                      &                      &                       & 0.8863 & 0.8035 & 0.9256 & 0.0184  \\
2 &  \Checkmark & \Checkmark &                       &                       & 0.9072 & 0.8260 & 0.9465 & 0.0113  \\
3 &  \Checkmark &  \Checkmark & \Checkmark  &                      & 0.9146 & 0.8361 & 0.9577 & 0.0095 \\
4 &  \Checkmark &  \Checkmark &                      & \Checkmark  & 0.9267 & 0.8554 & 0.9654 & 0.0078 \\

\hline
5 &  \Checkmark  & \Checkmark & \Checkmark  & \Checkmark & \textbf{0.9383} & \textbf{0.8740} & \textbf{0.9717} & \textbf{0.0061}   \\
\toprule
\end{tabular}
\end{table}

\subsection{Ablation Studies}
\label{Ablation Studies}
We conduct ablation studies on the EORSSD dataset to evaluate the contribution of each component of our TSCNet.
Concretely, we remove all TSCMs in our TSCNet, and directly connect $\boldsymbol{f}_\textrm{b}^{i}$ to the corresponding SP block, providing the baseline model.
As shown in Tab.~\ref{Ablation_component}, the performance of the baseline model drops significantly, with an average decrease of 5.62\%, which means our TSCMs are very effective.
The TSCM consists of three units of PAU and parallel TRU and RIU.
In the following, we gradually add these units to the baseline model.
First, we add PAU into the baseline model, resulting in an average performance improvement of 2.14\%.
Then, we continue to add TRU to the second variant, thereby increasing average performance by 0.96\% compared to the No.2 variant.
Meanwhile, we also add RIU to the second variant, increasing average performance by 2.26\% compared to the No.2 variant.
Finally, we add all three units to the baseline to achieve our complete TSCNet.
Thanks to the perfect collaboration between these three units, our complete TSCNet achieves satisfactory performance.

\section{Conclusion}
\label{sec:con}
In this brief, we make an attempt to investigate the collaboration of texture cues and semantic cues for ORSI-SOD, and propose a concise yet effective TSCNet.
We implement our TSCNet on the effective encoder-decoder structure, and integrate a useful TSCM into this structure.
In the TSCM, we first adopt semantic features to anchor the position of salient regions through joint channel attention and spatial attention.
Then, we use texture features to assist the super-resolution of current features through an efficient variant of the self-attention mechanism.
Meanwhile, we perform multi-scale perception and region-level interaction establishment to reconstruct the texture of features from a self-learning perspective.
The close collaboration of all components enables our TSCNet to not only accurately locate salient objects, but also sharpen their details.
Performance comparison and ablation studies demonstrate that our idea is effective, and our TSCNet exhibits competitive performance.



\ifCLASSOPTIONcaptionsoff
  \newpage
\fi

\bibliographystyle{IEEEtran}
\bibliography{TSCNetRef.bib}

%



%

\end{document}